\documentclass[
]{ceurart}

\usepackage{listings}
\usepackage{algorithm}
\usepackage{algpseudocode}
\begin{document}

\copyrightyear{2024}
\copyrightclause{Copyright for this paper by its authors.
  Use permitted under Creative Commons License Attribution 4.0
  International (CC BY 4.0).}

\conference{CREAI 2024: International Workshop on Artificial Intelligence and Creativity, ECAI, 2024, Santiago de Compostela, Spain}

\title{Collaborative Comic Generation: Integrating Visual Narrative Theories with AI Models for Enhanced Creativity}


\author{Yi-Chun Chen}[%
orcid=0009-0003-4035-9894,
email=ychen74@ncsu.edu,
url=https://sites.google.com/ncsu.edu/rimichen/home,
]

\author{Arnav Jhala}[%
email=ahjhala@ncsu.edu,
url=https://www.csc.ncsu.edu/people/ahjhala,
]
\address{North Carolina State University, Computer Science, Raleigh, NC 27606, USA}



\begin{abstract}
This study presents a theory-inspired visual narrative generative system that integrates conceptual principles—comic authoring idioms—with generative and language models to enhance the comic creation process. Our system combines human creativity with AI models to support parts of the generative process, providing a collaborative platform for creating comic content. These comic-authoring idioms, derived from prior human-created image sequences, serve as guidelines for crafting and refining storytelling. The system translates these principles into system layers that facilitate the creation of comics through sequential decision-making, addressing narrative elements such as panel composition, story tension changes, and panel transitions. Key contributions include the integration of machine learning models into the human-AI cooperative comic generation process, the deployment of abstract narrative theories into AI-driven comic creation, and a customizable tool for narrative-driven image sequences. This approach improves narrative elements in generated image sequences and brings engagement of human creativity in an AI-generative process of comics. We open-source the code at https://github.com/RimiChen/Collaborative\_Comic\_Generation.

\end{abstract}


\maketitle

\section{Introduction}

Despite their various names, comics, manga, and visual stories represent a dominant form of storytelling that spans cultures and age groups. Authors combine their creative storytelling ideas with textual expressions and graphical representations to convey intricate narratives through multi-modal panel sequences. Recently, generative AI, a trending topic in AI creativity, has explored the automatic generation of narratives and their synthesis with visual representations, simulating an activity traditionally rooted in human creativity. These studies lead to an exciting research question: how can AI collaborate with human creativity to create image sequences, such as comics and visual stories?

Nowadays, most generative AI models handle almost the entire process of creating image sequences, leaving little room for human authors to modify the details during generation. Traditionally, however, creating visual stories or comics relies heavily on authors' familiarity with narrative idioms \cite{cohn2014grammar}, principles of visual storytelling \cite{mccloud1993understanding}, and their skills in translating narrative content into visual representations \cite{martens2020visual}.

We propose a solution that balances these approaches, allowing AI models to assist in generating visual stories while also providing space for human creativity. Narrative theories and idioms are important tools for authors, helping them convey stories with clarity and making visual narratives more engaging for readers. These elements, derived from existing comics, serve as underlying schemes. We aim to leverage these narrative elements by integrating them into the generative process, enhancing the results.
Current studies on generating image sequences or visual storytelling include text-to-image synthesis \cite{copilot2023}, character consistency algorithms \cite{zhou2024storydiffusion}, and narrative structuring models \cite{jing2015content, gunasekara2024generate}. Although most synthesis methods allow users to provide some input as seeds, such as captions or base images, authors have limited flexibility in the creative process. Furthermore, while narrative idioms like grammar and patterns of visual storytelling are widely discussed in the analysis of image sequences, they are rarely incorporated into AI-driven comic generation. Even in studies that integrated rule-based methodologies to allow narrative theories to influence visual storytelling \cite{alves2007comics2d, martens2016discourse, martens2016generating, chen2023customizable}, authors' creativity remained largely excluded from the generative process.

Building on prior studies, we propose a comic-generating system that integrates a human-in-the-loop process, enabling authors to customize the generation by leveraging multiple AI models. Our system combines AI-driven narrative development and art generation, reducing the effort needed to create dynamic visuals. A sentiment analysis model also guides the story arc, serving as a plotline framework.

Second, our system dynamically applies narrative idioms during the generation process. The layer-wised customizations allow authors to edit the image sequence iteratively, selecting different rules to refine the results. This human-in-the-loop process ensures that authors can adjust based on the generated image sequences. Third, rather than treating the comic panel as a single entity, we decompose it into multiple layers: background, foreground, compositional layer, and symbol layer. This approach enables authors to customize specific details without altering the entire panel.

Finally, our system provides a Graphical User Interface (GUI) and an Application Programming Interface (API). These interfaces allow authors to extend and customize the comic generation process, creating engaging and varied content. Additionally, the system supports testing various machine learning models in comic generation. This approach balances AI and human creativity, making the comic creation process more flexible and efficient.

\section{Methodology}
\begin{figure*}[ht]
    \centering
    \includegraphics[width=0.6\linewidth]{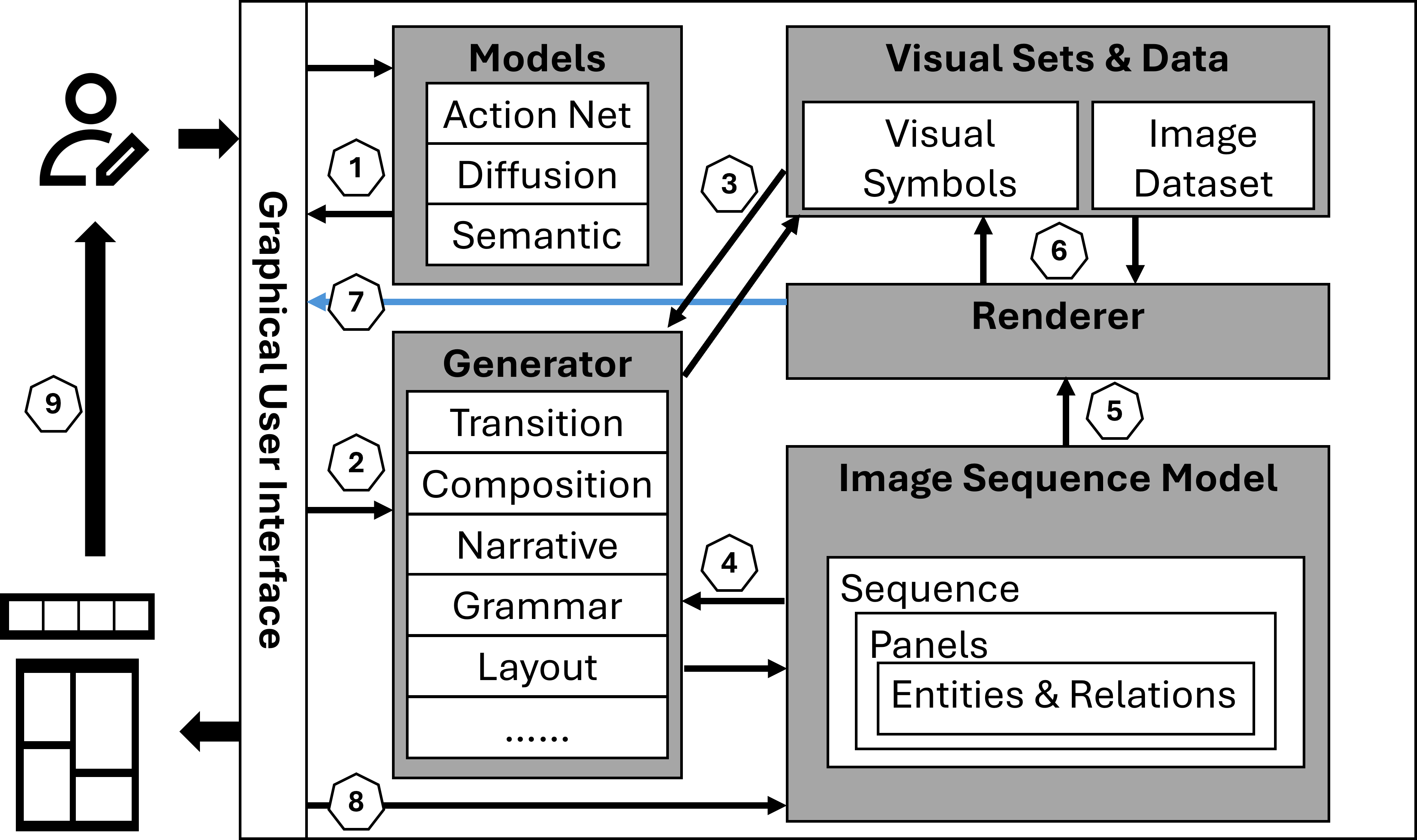}
    \caption{Overall system architecture and the author-in-loop workflow.}
    \label{OverallWorkflow}
\end{figure*}

The system is structured to a human-in-loop workflow as Figure \ref{OverallWorkflow}.

The system comprises six modules: the graphical interface, a container of models, the pool of visual sets and data, an image sequence model, a generator, and a renderer.

\begin{itemize}
    \item \textbf{Graphical Interface}: This is the primary means of interaction for human authors. It allows them to input base images, launch and apply ML models to the current image sequence, import scripts for customized editing layers, and perform simple operations like selection and dragging to modify images.
    \item \textbf{Container of Models}: This module houses various ML models that can be applied during generation.
    \item \textbf{Pool of Visual Sets and Data}: This repository contains visual elements and data sets the system can draw upon to create the image sequences.
    \item \textbf{Image Sequence Model}: This model organizes the graph model for linking the elements of the image sequence, including sequence information, panel transitions, panel content, characters, and narrative elements.
    \item \textbf{Generator}: This component integrates the customized editing layers into a pipeline that iteratively edits the image sequence by applying the narrative goals of each layer.
    \item \textbf{Renderer}: This module finalizes the visual representation of the generated sequences, ensuring they are ready for presentation or further editing.
\end{itemize}

Due to the complexity of applying ML models, some customizations of the generating process rely on scripts that deploy API functions across different system modules. The process typically follows nine steps between these modules:

\begin{enumerate}
    \item \textbf{Register Models}: After receiving user scripts as input, the \textit{Container of Models} registers all imported ML models and initializes them.

    \item \textbf{Register Editing Layers}: This step links the interface buttons with class scripts, dynamically imports user-defined classes, and executes them upon button click.
    \item \textbf{Retrieve Data}: If user-customized scripts include functions to retrieve information, the \textit{Generator} communicates with the \textit{Pool of Visual Sets and Data} to obtain necessary data.

    \item \textbf{Apply Editing}: The customized editing results reflect the corresponding changes to the data nodes in the \textit{Image Sequence Model}.

    \item \textbf{Compose Visual Panels}: After updating the \textit{Image Sequence Model} with the desired changes, the \textit{Renderer} is triggered to start composing the visual results of comic panels. The updated graph model will also be passed to the \textit{Renderer}.

    \item \textbf{Map Semantic to Visual}: The \textit{Renderer} maps the data nodes to visual elements in the \textit{Pool of Visual Sets and Data} and forms the multi-layered panels. These layers include background, foreground, composition, symbol, and any other user-customized layers.

    \item \textbf{Render}: The multi-layered image panel is passed to the interface to support user selection and drag functions.

    \item \textbf{Modify Results}: The interface inputs user interactions to update the \textit{Image Sequence Model} and repeats the rendering steps.

    \item \textbf{Get Result}: The generated and modified image sequence is presented on the interface for users to view.
\end{enumerate}

\subsection{Build-in and Extendable Elements:}
To demonstrate the system's capabilities, we incorporated a built-in model, the action causal graph, and a visual symbol set as the default sources to support the plotline of the visual representations. These components are extendable through the APIs of the \textit{Container of Models} and the \textit{Pool of Visual Sets and Data}, respectively. Detailed documentation will be provided in the API subsections.
\subsubsection{Common Symbols}

In human-created comics, authors often use abstract symbols to visualize ideas such as atmosphere, motion, and emotions, thereby exaggerating characters' reactions. These symbols include emojis to emphasize emotions, speed lines to show movement, explosion shapes to represent collisions, cross shapes to denote anger, and many others. Table \ref{symbol} presents examples from the Manga109 dataset, a collection of 109 Japanese manga titles published in commercial magazines \cite{narita2017sketch, matsui2017sketch}. These examples serve as references for the symbols or emojis used in our default set.

\begin{table}
\centering
\begin{tabular}{|p{1.4cm}|p{1.4cm}|p{1.4cm}|p{1.4cm}|p{1.4cm}|p{1.4cm}|p{1.4cm}|p{1.4cm}|}
    \hline
    Anger & Quick moving & Slow moving & Anxious & Collision & Relieved & Shock & Big shock\\
    \hline
    \includegraphics[width = \linewidth]{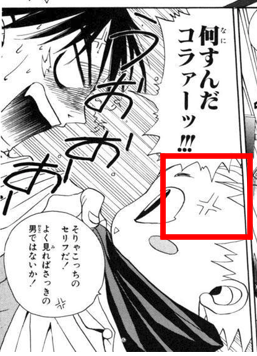}
    &\includegraphics[width = \linewidth]{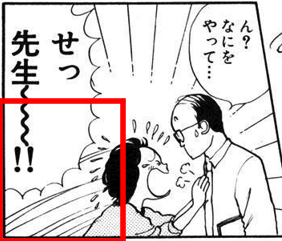}
    &\includegraphics[width = \linewidth]{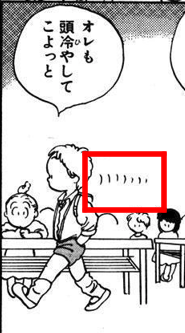}
    &\includegraphics[width = \linewidth]{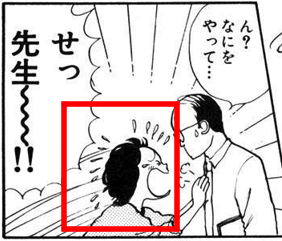}
    &\includegraphics[width = \linewidth]{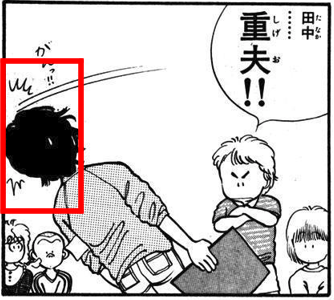}
    &\includegraphics[width = \linewidth]{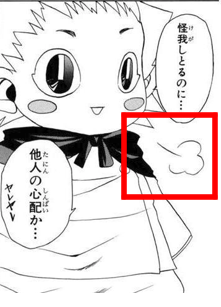}
    &\includegraphics[width = \linewidth]{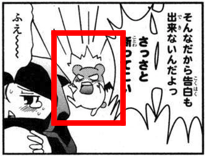}
    &\includegraphics[width = \linewidth]{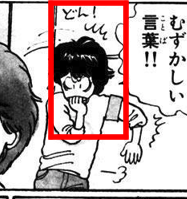}\\
    \hline     
\end{tabular}
    \caption{
    \label{symbol} \small{Examples of metaphor symbols of action and characters' emotions from Manga109.AisazuNihaIrarenai© Yoshimasako, AkkeraKanjinchou© Kobayashiyuki,Akuhamu© Araisatoshi}
    }
\end{table}
\subsubsection{Action Causal Network}
The \textit{Action Causal Graph} is a directed graph model where each node represents an action that a character might perform, and the links indicate the causal relationships between actions and possible reactions. For example, "Fall" links to "Fly," "Jump," and "Run," while "Dizzy," "Collide," and "Hit" link to "Run." Consecutive nodes form action pools for plot planning. The current version of the default action set built into the system includes a small group of common daily actions. The detailed method for expanding the graph and its scalability will be discussed in subsequent subsections.


    
\subsection{AI-Assisted Editing Layers and Narrative Theories:}
We implement three ML-driven editing layers in the generator to demonstrate the system's image sequence pipeline. The first layer uses a stable diffusion model for modifying visual elements, while the second layer combines the PAD emotion model with a semantic analysis language model to guide plotline decisions.
\subsubsection{Diffusion Model for Visual Elements}
We employed a pre-trained stable diffusion model developed by the CompVis group to implement one of the editing layers \cite{rombach2021highresolution, rombach2022stablediffusion}. This high-resolution image synthesis model transforms input text descriptions into detailed and coherent images based on latent diffusion. The model allows users to adjust visual elements like characters or scenes.

The generated panel sequences are rendered through our pipeline, where information is first updated in the Image Sequence Model before being rendered. Any changes in the visual representation of an entity will be reflected in its semantic nodes, ensuring character consistency throughout the panel sequence. For example, if the diffusion model alters the visual representation of character\_x, this change will be mirrored in its semantic mapping. Similar rules apply to other visual elements. Furthermore, our system supports multi-layer rendering, dividing panel images into background, foreground, compositional, and symbol layers. This feature enables partial redrawing of the generated panel, ensuring that changes applied to one entity do not interfere with others.

\subsubsection{Plotline and Narrative Theories}
To form a simple plotline, we incorporate narrative idioms and theories. This editing layer aims to generate content for the image sequence according to a specific narrative arc. We use Cohn's narrative grammar \cite{cohn2013visual, cohn2015analyze, cohn2016visual} to estimate the narrative arc, as the grammar categories indicate plot changes throughout the comic sequence. After establishing the narrative arc, we target the characters' actions to predict story tension.


We use the arousal level concept from the PAD emotion model \cite{mehrabian1974basic, russell1981affective, russell1980circumplex} and sentiment labels from the language model to predict arousal scores for character actions, where higher scores indicate greater story tension. Differences in arousal scores between consecutive actions (based on our action causal graph) form a probability distribution for subsequent actions. This mapping enables the editing layer to select actions probabilistically while maintaining narrative arc alignment. Detailed explanations of each component are provided in the following subsections.\\

\textbf{Narrative Grammar}\\

We formalize the plot generation of new comic sequences using Cohn's Visual Narrative Grammar (VNG). Starting with the narrative structure to determine the content's global reasoning, we adopt Cohn's theory, which proposes that coherent image sequences follow a grammar, organizing their global structure into five categories. 

\begin{itemize}
    \item \textbf{Establisher(E)}: Sets the objects and scenes without involving any action.
    \item \textbf{Initial(I)}: Marks the beginning of a story arc—the starting point of a sequence of actions or events.
    \item \textbf{Prolongation(L)}: Represents the middle state of the story arc, extending an action.
    \item \textbf{Peak(P)}: Indicates the highest story tension—the climax of an action.
    \item \textbf{Release(R)}: Releases the tension—the outcome or result of an action.

\end{itemize}
The five categories form basic phases through linear ordering:\\
	\textbf{Phase (Establisher) - Initial(Prolongation) - Peak - (Release)}\\

The use of parentheses indicates that categories are optional when forming phases. The categories have different levels of importance, ranked from highest to lowest as follows: Peak, Initial, Release, Establisher, and Prolongation. Additionally, more complex combinatorial structures can be created through the conjunction of embeddings. Our editing layer generates the narrative structure using center-embedding, expanding a new tree structure by replacing a single category with a phase.

Table \ref{grammar} provides an example of comic sequences (illustrated with simplified icons) following this grammar structure. In the same scene, the circle character appears in the first panel and then begins performing actions in the subsequent panels. In the fourth panel, a significant event occurs, creating a small climax, which leads to the resolution in the final panel. This sequence follows the narrative structure \textit{E-I-L-P-R}.



\begin{table}
\centering
\begin{tabular}{|p{1.5cm}|p{1.5cm}|p{1.5cm}|p{1.5cm}|p{1.5cm}|}
    \hline
    E & I& L& P & R\\
    \hline
    \includegraphics[width = \linewidth]{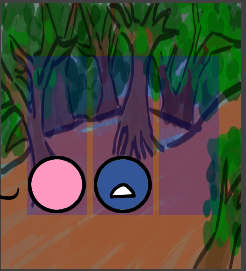}
    &\includegraphics[width = \linewidth]{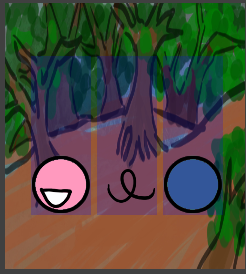}
    &\includegraphics[width = \linewidth]{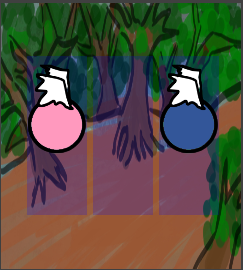}
    &\includegraphics[width = \linewidth]{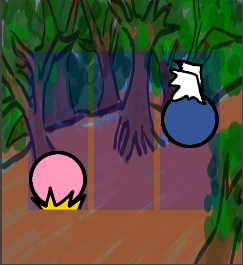}
    &\includegraphics[width = \linewidth]{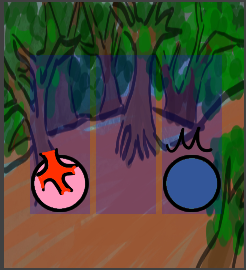}\\
    \hline     
\end{tabular}
    \caption{
    \label{grammar} \small{Examples of a comic sequence that followed grammar categories.}
    }
\end{table}

The implementation follows the algorithm flow below:

\begin{algorithm}
\caption{Grammar Structure}\label{grammar_algo}
\begin{algorithmic}[1]
\Procedure{Grammer Structure Script}{}
\State $P$ $\gets$ input the current panel sequence.
\State $VNG$ $\gets$ Create object list for VNG basic phases.
\State $S$ $\gets$ Expend a center-embedded tree with $VNG$, get the reference narrative structure.
\If   $Length(P)$ $\neq$ $Length(S)$
    \State Add or Subtract empty panels from $P$  
\EndIf
\State \emph{loop:}
    Assign each phases in $VNG$ to $P$
\State \Return $P$

\EndProcedure
\end{algorithmic}
\end{algorithm}

The input is a comic sequence, either a generated result or an empty sequence with a certain length. Then, the editing layer creates an object dictionary for grammar phases and then expands the tree structure. The structure then decides the narrative arc of the comic sequence. It assigned a grammar phase to a sequence, and we then formed the possible narrative arc based on the structure.\\

\textbf{Narrative Arc}\\

The importance ranking of narrative grammar categories mirrors the narrative arc, reflecting a story's progression from a calm beginning through a peak of tension in the middle to conflict resolution at the end. The editing layer projects these categories onto a curve that illustrates changes in story tension.

\begin{algorithm}
\caption{Narrative Arc}\label{narrative_arc}
\begin{algorithmic}[1]
\Procedure{Mapping Narrative Arc with Narrative Structure}{}
\State $P$ $\gets$ input the current panel sequence.
\State $REF$ $\gets$ create value dictionary with VNG phases with assigned tension scores.
\State $REF = \{ E:0 I: 2 L: 4 P: 6 R:2\}$
\If {panels in $P$ have assigned grammar phase}
\State \emph{loop:} over $P$, apply $REF$ with the grammar phase 
\Else {use default narrative arc scores}
\EndIf
\State \Return $P$

\EndProcedure
\end{algorithmic}
\end{algorithm}

To capture the abstract concept of tension, we assigned each grammar phase a score between one and ten based on its narrative function. For example, the Peak(P) phase has the highest tension, while the Release(R) phase somewhat alleviates the tension. We begin by creating a value dictionary for each grammar phase. The next step is to map these phases to their respective scores and generate the curve of the narrative arc.\\

\textbf{PAD Emotion State Model and Semantic Analysis}\\

The PAD emotion state model, developed by Albert Mehrabian and James A. Russell, describes emotions through three dimensions: Pleasure, Arousal, and Dominance. This model quantifies emotional states, making it valuable in psychology, user experience design, and AI. Specifically, the Arousal dimension measures how energized or calm one feels, reflecting the activation level of an emotion. We adapt this concept to model narrative momentum—story tension.

Considering the image sequence-generating system's future expansion and possible integration with a more extended narrative, we employed a sentiment analysis model for sentences--Roberta base model for emotion classification, fine-tuned on the GoEmotions dataset \cite{Lowe2021}. We then project the sentiment labels in the RoBERTa model to the PAD model's emotion labels, dividing the emotions into high, medium, and low arousal levels. By computing the feature vectors through the BERT model \cite{devlin2019bert} of the two label sets, we can measure the Euclidian distance between the labels, estimating the possible arousal level scores for the sentiment labels. The mapping process follows the flow below: 

\begin{algorithm}
\caption{Mapping Sentiment Labels with Emotion Labels}\label{mapping_labels}
\begin{algorithmic}[1]
\Procedure{Estimating Arousal Levels}{}
\State $E$ $\gets$ emotion labels from the PAD emotion state model, where [high, medium, low] arousal level maps to [1, 0, -1].
\State $S$ $\gets$ sentiment class labels from the RoBERTa model.
\State $Sdis$ $\gets$ Distance matrix for distance between labels in $S$ and labels in $E$.
\State \emph{loop:} over $S$, compute the distance to each element in $E$, and get $Sdis$ 
\State \emph{loop:} over each row in $Sdis$, flat the vector except the minimum two elements in the row.
\State \emph{loop:} over each row in $Sdis$, use $Sdis/ sum(Sdis)$ as the weight multiply with $E$ to get $S\_arousal$.
\State normalize $S\_arousal$ to $[-1, 1]$ 
\State \Return $S\_arousal$

\EndProcedure
\end{algorithmic}
\end{algorithm}

\begin{figure*}[ht]
    \centering
    \includegraphics[width=0.7\linewidth]{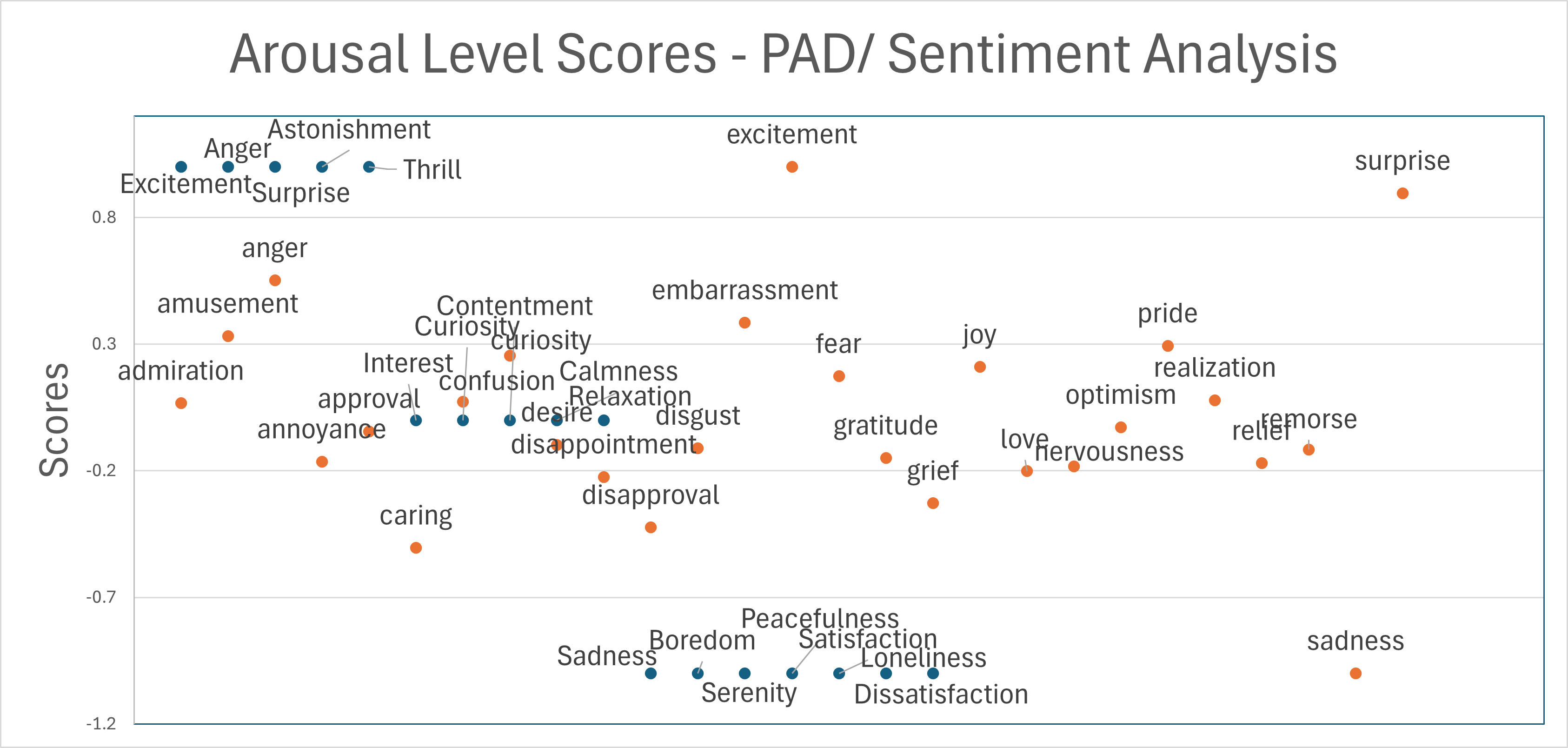}
    \caption{The arousal level scores are estimated using label set mapping.}
    \label{emotionlabels}
\end{figure*}

Figure \ref{emotionlabels} shows the results, where the blue points represent the emotion labels from the PAD model, and the orange points represent the sentiment classes in the language model.

The sentiment analysis model predicts probabilities across various sentiment classes when using a sentence or word as input. We use these probabilities as weights and then measure the distance between the input action and the sentiment classes, predicting the possible score of the input action. This score indicates the story's tension. We further compute the slope between two consecutive actions and divide it by the sum of slopes for all subsequent actions. Finally, we normalize this value to determine the probability of the following action.\\

\textbf{Action Mapping} \\
By mapping arousal scores with actions, the editing layer integrates the narrative arc and actions by referencing these scores. Using the curve from the narrative arc, the layer calculates changes along the curve and selects actions that best fit these changes. Additionally, it sets a likelihood tolerance with the probability of actions, expanding potential narrative diversity. The process is described below:


\begin{algorithm}
\caption{Narrative Arc Mapping}\label{score_mapping}
\begin{algorithmic}[1]
\Procedure{Mapping Actions to Fit Narrative Arc }{}
\State $P$ $\gets$ input the current panel sequence.
\State $NET$ $\gets$ get the action causal graph network.
\State $ACT$ $\gets$ creates a value dictionary for actions according to the arousal scores.
\State $ARC$ $\gets$ get referenced Narrative Arc.
\State \emph{loop:} over $P$, check the characters' actions in the next panel and compute the score difference, according to $ARC$
\State Select actions in likelihood from $ACT$ and $NET$
\State \emph{loop:} revise action selection other panels according to $NET$
\State \Return $P$
\EndProcedure
\end{algorithmic}
\end{algorithm}

\subsubsection{Panel Relations}




Panel transitions refer to the changes in content between consecutive panels. McCloud proposed six categories of transition types to model various aspects of content change\cite{mccloud1993understanding}, while Cohn introduced conjunction schemes to capture more complex panel transitions \cite{cohn2013visual}. Our system combines these theories to modify the visual composition of comic panels—how elements are arranged—and reflect content changes according to the transition types. The narrative goal of this editing layer is to arrange the panel transitions in the generated comic sequence to create dynamic viewport changes and increase tension.

Here is how we map the transitions with panel content changes:
\begin{itemize}
    \item \textbf{Action}: McCloud's action-to-action transition indicates changes in actions between consecutive panels. We use this transition to guide the selection of different character actions in the next panel.
    \item \textbf{Scene}: McCloud's scene-to-scene transition indicates changes in scenes between consecutive panels. We use this transition to guide the change of location where the character's action occurs in the next panel.
    \item \textbf{Object}: The object-to-object transition indicates a shift in focus from one object to another. We use this transition to guide the focus to different objects.
    \item \textbf{Addition}: Cohn's additive conjunction involves panels that add information or detail to the ongoing story. We use this transition to introduce new objects into the panels.
    \item \textbf{Alternation}: Cohn's alternating conjunction presents alternative scenarios or actions, offering possibilities within the narrative. Compared to scene transitions, we use this transition to guide panels to alter most elements while maintaining consistent characters.
\end{itemize}

In our system's current version, some complex transitions or conjunctions are not yet supported but can be added through customizations. For example, the Temporal Conjunction depicts the progression of time, a rather abstract concept for visual representation. Another example is the Contrasting Conjunction, which depicts opposing ideas, actions, and emotions; achieving this requires deep semantic analysis of the narrative. However, we have integrated causal conjunctions to some extent by using action causal graphs.

\section{Usage Explanations and Showcases}

This section introduces our system's application programming interface (API), and graphical user interface (GUI).
The API functions are in Table \ref{tab:api}. The system was constructed using several core classes, with the \textit{Parameter}, \textit{Layer}, and \textit{AttributeNode} classes being the most crucial. The \textit{Parameter} class manages all the registers of models and scripts. The \textit{AttributeNode} class serves as the fundamental component of the graph model, representing the entire sequence. The \textit{Layer} class is the parent class for all editing layer scripts. To execute modifications, the \textit{apply} methods within the \textit{Layer} class must be overridden and then executed by the Generator module.

\begin{table}[htbp]
\centering
\begin{tabular}{| m{2.5cm} | m{3.5cm} | m{5.5cm} | m{2cm} |}
\hline
\textbf{Function Name} & \textbf{Parameters} & \textbf{Description} & \textbf{Return Type} \\ 
\hline
\texttt{Parameter()} & 
Basic parameter for GUI settings
\begin{itemize}
    \item  win\_w, win\_h
    \item  menu\_w
    \item sequence\_len 
\end{itemize} & The class that registers and manages all the modules. & None \\ 
\hline
\texttt{addModel} &
Selected ML model class
\begin{itemize}
    \item module\_name
\end{itemize}
& Register the ML models for use in the generating process.
& None \\ 
\hline
\texttt{importModels()} &
None
& Import all the registered models.
& None \\ 
\hline
\texttt{Models()} 
& 
None
& Initialize the Container of Models and initialize all the registered models.  
& None \\ 
\hline
\texttt{addModule()} &
Customized editing layer class
\begin{itemize}
    \item layer\_name
\end{itemize}
& Register script for a new editing layer.
& None \\ 
\hline
\texttt{importModules()} &
None
& Import all the registered editing layers.
& None \\ 
\hline
\texttt{Sequence()} & 
Initialized Parameter object and attribute node type
\begin{itemize}
    \item attribute\_type
    \item Parameter() 
\end{itemize} 
& Inherit from the AttributeNode class and initialize the root of the graph model for the Image Sequence Model.  
& None \\ 
\hline
\texttt{Generator()} 
& 
None
& Initialize the Generator Module.  
& None \\ 
\hline
\texttt{addVisuals()} & 
Name of target visual set, file path, or folder path for image input
\begin{itemize}
    \item set\_name
    \item path
\end{itemize} 
& Expand or create the visual set with the assigned name.  
& None \\ 
\hline
\texttt{GUIInterface()} & 
Initialized Parameter object
\begin{itemize}
    \item Parameter() 
\end{itemize} 
& Initilize the GUI and the Renderer.  
& None \\ 
\hline
\texttt{Layer()} & 
None
& The Parent class of all the editing layers.  
& None \\ 
\hline
\texttt{Layer.apply()} & 
The graph model of a generated sequence
\begin{itemize}
    \item Sequence()
\end{itemize} 
& Apply the modifications to the comic sequence and return the results.   
& Sequence() \\ 
\hline
\texttt{AttributeNode()} & 
The attribute type and the initialized Parameter class
\begin{itemize}
    \item attribute\_type
    \item Parameter()
\end{itemize} 
& The Parent class of all the attribute nodes, including Sequence, Panel, Character, etc. It is the Node class for the graph model in the Image Sequence Model.
& None \\ 
\hline
\texttt{addAttribute()} & 
The parent node's name, attribute type, and the child node (self). 
\begin{itemize}
    \item parent\_node
    \item attribute\_type
    \item AttributeNode()
\end{itemize} 
& Add an attribute node to the graph model as a child node of the assigned parent node.   
& None\\ 
\hline

\end{tabular}
\caption{API Functions}
\label{tab:api}
\end{table}

\begin{figure*}[htbp]
    \centering
    \includegraphics[height=0.3\linewidth]{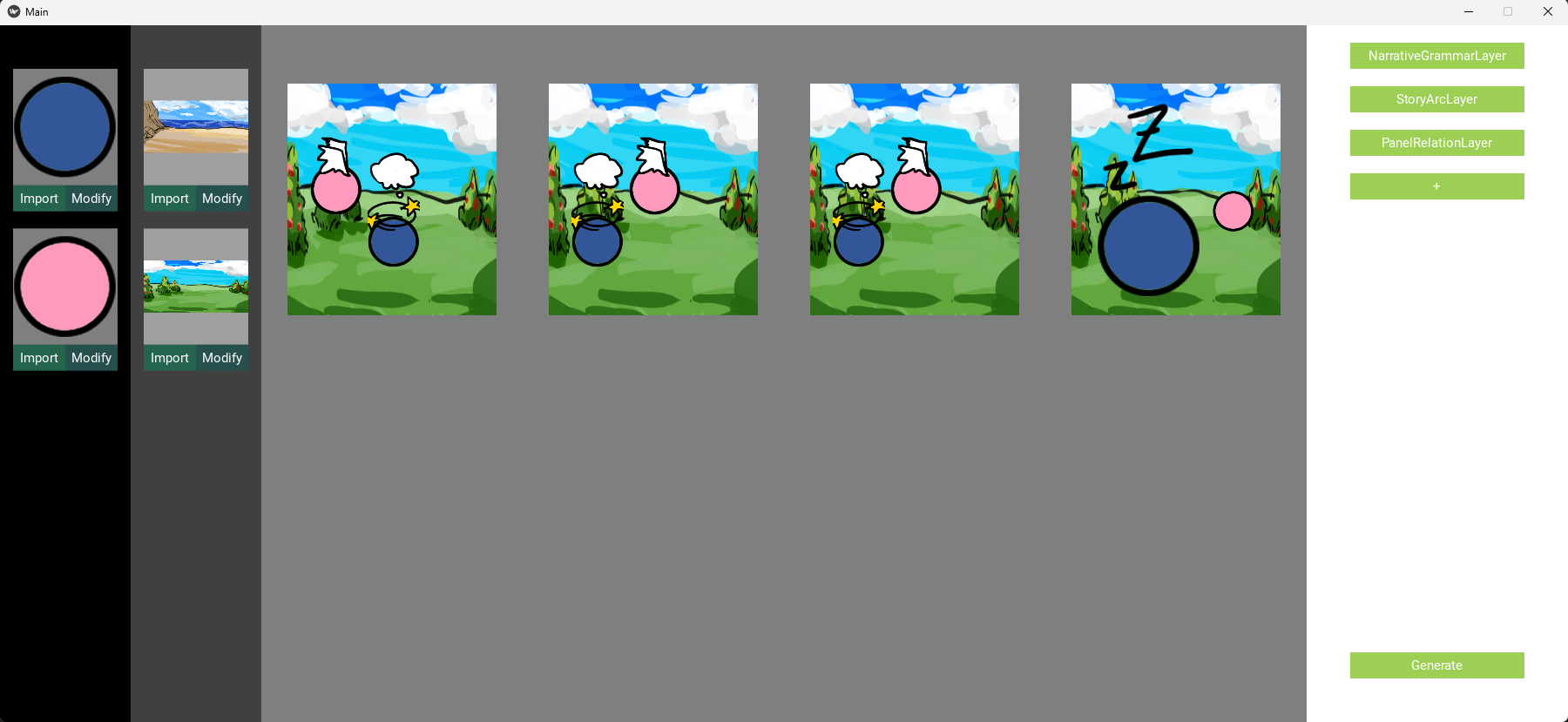}
    \caption{The graphical user interface of the generating system.}
    \label{comicGenerator}
\end{figure*}

The GUI screenshot is Figure \ref{comicGenerator}. The two columns on the left display the image inputs for the character and scene of the comic sequence. Users can generate these images using the diffusion model or import them from their work. The column on the right side contains buttons that link to the imported scripts, triggering functions to apply modifications and generate results. The large area in the middle shows the currently generated result and allows users to select comic panels and the elements within them.

\vspace{-0.5cm}
\subsection{Showcases}

Table \ref{modification} compares two sets of generated results, one before and one after applying the changes implemented in the current version. The first set, generated with user-imported images and all editing layers turned off, features content with default panel composition and randomly chosen characters' actions. After partial redrawing with the stable diffusion model, the second set demonstrates the results, followed by an editing layer that uses the sentiment language model to achieve narrative planning. It includes a partially redrawn icon and scene using a diffusion model from an Einstein head icon and a WindowsXP desktop photo as inputs. In addition, based on the action causal network, it shows a short narrative in which the two characters ate apples and felt dizzy after eating, then shocked and finally rested in the garden. 

\begin{table}
\centering
\begin{tabular}{|p{1.5cm}|p{2cm}|p{2cm}|p{2cm}|p{2cm}|}
    \hline
    Set\# &  & & & \\

    \hline
    1
    &\includegraphics[width = \linewidth]{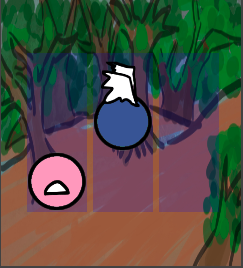}
    &\includegraphics[width = \linewidth]{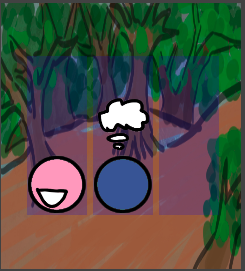}
    &\includegraphics[width = \linewidth]{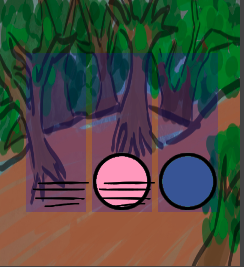}
    &\includegraphics[width = \linewidth]{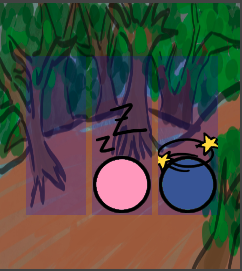}
   \\
    \hline     
    2
    &\includegraphics[width = \linewidth]{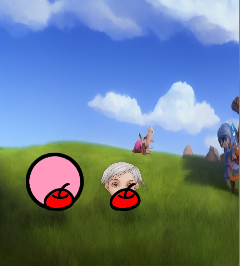}
    &\includegraphics[width = \linewidth]{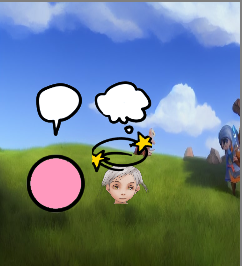}
    &\includegraphics[width = \linewidth]{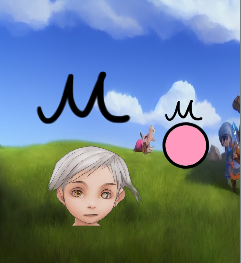}
    &\includegraphics[width = \linewidth]{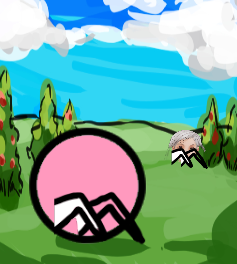}
   \\
    \hline     

\end{tabular}
    \caption{
    \label{modification} \small{Examples of before and after applying the changes.}
    }
\end{table}


\subsection{Data Availability}
The data and code used in this study are openly available in a GitHub repository. The repository includes the raw data, processed data, and all scripts necessary to reproduce the analyses presented in this paper. You can access the repository at https://github.com/RimiChen/Collaborative\_Comic\_Generation. The repository is licensed under MIT License, allowing for reuse and modification with appropriate attribution.



\section{Conclusion and Future Work}

This paper presents an extensible system for generating comic-style visual narratives, integrating narrative theory with human-AI collaboration. We address challenges in visual modification and plotline planning, balancing automation with user control for customization.


While the current system effectively integrates abstract narrative theories, it has limitations in coordinating visual components and diversifying narratives. Separating visual layers, though beneficial for user modifications, reduces scene and character interaction, limiting cohesive artwork and rich actions. Additionally, using symbols to represent actions restricts narrative diversity, requiring user customization to expand content. Integrating advanced language models could enhance narrative richness

Future work will focus on refining the integration of narrative and visual components and exploring advanced models to enhance the system's capability to handle diverse narratives. We also plan to conduct user studies to evaluate effectiveness and usability across different groups. In conclusion, our work advances human-AI cooperative visual narrative generation, offering a versatile platform for creating engaging experiences.

\bibliography{Reference}


\end{document}